\documentclass[lettersize,journal]{IEEEtran}
\usepackage{amsmath,amsfonts}
\usepackage{algorithmic}
\usepackage{algorithm}
\usepackage{array}
\usepackage[caption=false,font=normalsize,labelfont=sf,textfont=sf]{subfig}
\usepackage{textcomp}
\usepackage{stfloats}
\usepackage{url}
\usepackage{hyperref}
\usepackage{verbatim}
\usepackage{graphicx}
\usepackage{cite}
\usepackage{makecell, rotating, multirow, diagbox}
\usepackage{arydshln}
\usepackage{color, colortbl}
\usepackage[table, dvipsnames]{xcolor}
\def\eg{{\it e.g.}}
\def\ie{{\it i.e.}}
\def\etc{{\it etc.}}
\def\etal{{\it et al.}}
\usepackage{pifont}
\begin{document}

\title{GLFF: Global and Local Feature Fusion for AI-synthesized Image Detection
\thanks{This work is supported by the US Defense Advanced Research Projects Agency (DARPA) Semantic Forensic (SemaFor) program, under Contract No. HR001120C0123.}
\thanks{Y. Ju, S. Jia, J. Cai, and S. Lyu are with the Department of Computer Science and Engineering, University at Buffalo (UB), State University of New York, NY, USA. E-mail: (yanju, shanjia, jialingc, siweilyu)@buffalo.edu. 
H. Guan is with the National Institute of Standards and Technology (NIST), 100 Bureau Drive, Gaithersburg, MD 20899, USA. E-mail: haiying.guan@nist.gov.
  
Certain equipment, instruments, software, or materials are identified in this paper in order to specify the experimental procedure adequately.  Such identification is not intended to imply recommendation or endorsement of any product or service by NIST, nor is it intended to imply that the materials or equipment identified are necessarily the best available for the purpose.
  
Corresponding author: Siwei Lyu.}}

\author{Yan Ju, Shan Jia, Jialing Cai, Haiying Guan, Siwei Lyu,~\IEEEmembership{Fellow,~IEEE,}}
        


\maketitle


\begin{figure*}[!t] 
\centering 
\includegraphics[width=0.95\textwidth]{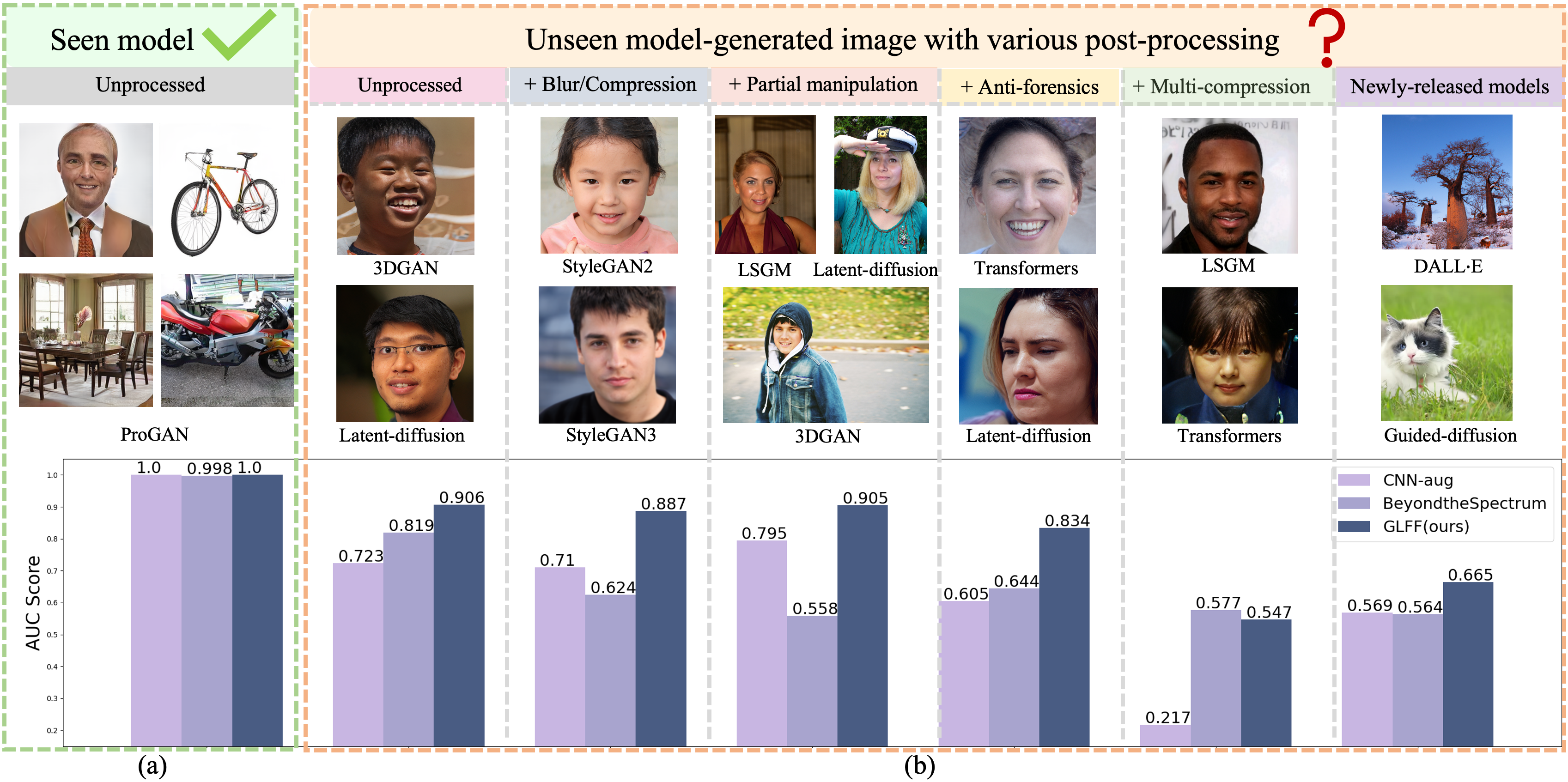}
\vspace{-0.3cm}
\caption{\textbf{ {Illustration of the unseen challenges to current synthetic image detectors. Note that existing state-of-the-art methods CNN-aug~\cite{wang2020cnn} and BeyondtheSpectrum~\cite{yang_ijcai21} achieve good performance on seen unprocessed images (as shown in (a)) but drop greatly on unseen post-processed images (as shown in (b)), while our method (GLFF) achieves good performance on both seen and unseen, and unprocessed and processed images. See texts for more details. This figure is better viewed in color.}}} 
\vspace{-0.4cm}
\label{fig:teaser1} 
\end{figure*}

\begin{abstract}
With the rapid development of deep generative models (such as Generative Adversarial Networks and Diffusion models), AI-synthesized images are now of such high quality that humans can hardly distinguish them from pristine ones. Although existing detection methods have shown high performance in specific evaluation settings, \eg, on images from seen models or on images without real-world post-processing, 
they tend to suffer serious performance degradation in real-world scenarios where testing images can be generated by more powerful generation models or combined with various post-processing operations. 
To address this issue, we propose a Global and Local Feature Fusion (GLFF) framework to learn rich and discriminative representations by combining multi-scale global features from the whole image with refined local features from informative patches for  {AI-synthesized image detection}. GLFF fuses information from two branches: the global branch to extract multi-scale semantic features and the local branch to select informative patches for detailed local artifacts extraction. 
Due to the lack of a  {synthesized image dataset} simulating real-world applications for evaluation, we further create a challenging fake image dataset, named DeepFakeFaceForensics ($DF^3$), which contains $6$ state-of-the-art generation models and a variety of post-processing techniques to approach the real-world scenarios. 
Experimental results demonstrate the superiority of our method to the state-of-the-art methods on the proposed $DF^3$ dataset and three other open-source datasets. 
\end{abstract}


\vspace{-0.2cm}
\section{Introduction}

\IEEEPARstart{T}{he} {creation of realistic synthetic images using deep learning models has seen rapid development with improved visual qualities and run-time efficiencies in recent years. Notable examples such as face images synthesized with the Generative Adversarial Networks~(GANs)~\cite{karras2020analyzing}, Diffusion models~\cite{ho2020denoising} have become challenging for humans to distinguish~\cite{korshunov2020deepfake} and eroded the trust in domains such as social media, politics, the military, geospatial intelligence, and cybersecurity. Malicious media generated by generative tools are major threats in impersonation and disinformation, and correspondingly, a variety of detection methods~\cite{lyu2020deepfake, wang2022gan} have been developed to expose them. }

Most existing detection methods employ deep neural networks to automatically extract discriminative features from face images 
and then feed them into a binary classifier to discriminate between the real and fake ones~\cite{wang2020cnn,zhao2021multi,gragnaniello2021gan,asnani2021reverse}. 
These detectors have achieved good performance on images from seen generation models, where the testing samples are generated using the same model structure or similar source data as the training samples. However, their performance tends to decrease considerably in open-world applications, where the testing images come from a different domain such as being generated by unknown models using unseen source data~\cite{ju2022fusing,hulzebosch2020detecting}. When facing such diverse and unknown testing samples, the features learned by current detection models can hardly conduct robust and accurate detection~\cite{hulzebosch2020detecting}. To improve the detection robustness and generalization ability, recent studies have explored different deep learning techniques to learn more robust and generic features, such as using limited sub-sampling architecture with contrastive learning~\cite{cozzolino2021towards}, transfer learning~\cite{jeon2020t}, incremental learning~\cite{marra2019incremental}, representation learning~\cite{kim2021fretal}, data augmentation~\cite{wang2020cnn}, spectrum analysis~\cite{frank2020leveraging}, and dual-color feature fusion~\cite{chen2021robust}.

Although these detectors have achieved promoted performance in detecting unseen synthesized images, their models are evaluated in specific evaluation settings, \ie, with initial generation models and limited post-processing operations. In real-world applications, the detection methods will inevitably face two kinds of challenges that may lead to performance degradation: 1) The quality of fake images has improved with refined global qualities and subtle artifacts in local regions due to the rapid development of generation models; 2) Well-designed post-processing operations are inevitable, such as image compression when laundering on social media, secondary manipulation, or well-designed adversarial noise for anti-forensics~\cite{ding2021anti}. 
Previous methods with limited feature representations, \eg, only utilizing global spatial features~\cite{cozzolino2021towards, wang2020cnn, kim2021fretal, chen2021robust} or focusing on local features\cite{cozzolino2017recasting, chai2020makes}, can hardly extract rich and strong representations that can generalize to powerful generation models or diverse post-processing operations in real-world images. We illustrate this finding in Fig.~\ref{fig:teaser1}, which shows that the high performance achieved in seen data domain drops significantly in detecting unseen and diverse post-processed images. 

To address this issue, we emphasize the importance of combining discriminative and complementary features inspired by the powerful feature learning strategies of previous fusion-based detection works~\cite{zhao2021multi, chen2021robust, zhang2021thinking}. Instead of purely focusing on feature fusion from the global domain, we propose a Global and Local Feature Fusion (GLFF) framework to fuse multi-scale global and local subtle features to learn more robust and general feature representation for challenging  {AI-synthesized image detection}. Specifically, our GLFF consists of two branches: the global branch combines low-level shallow features with high-level semantic features with a proposed Attention-based Multi-scale Feature Fusion~(AMSFF) module. Taking the fused multi-scale features as guidance, our local branch utilizes the Patch Selection Module~(PSM) to locate informative patches automatically and then extract local and subtle artifacts. The two-branch features are finally fused using an attention-based module and fed into a binary classifier to distinguish between real and fake images. 

Considering the lack of a comprehensive  {synthesized face dataset} to evaluate how detection methods will perform in a real-world scenario, we further present a challenging  {synthesized face image dataset}, $DF^3$, which combines state-of-the-art image generation models with a variety of post-processing operations to approach the real-world applications. Specifically, we include $6$ recent generation models, including GAN-based, Transformer-based, and diffusion-based generation models proposed in recent two years. Furthermore, we consider the following post-processing operations conducted on the generated face images: 1) common operations used on social media, \eg, JPEG compression and Gaussian blurring; 2) secondary manipulation with face blending, which blends the generated face into a real background to enhance the authenticity of the forged images; 3) several well-designed anti-forensics algorithms; 4) a novel multi-image compression to create the comprehensive and diverse dataset; 5) mixed operations by combining multiple operations. Overall, the $DF^3$ dataset contains approximately 46,400 generated face images.

Our main contributions can be summarized as follows:
\begin{itemize}

    \item We propose a novel detection framework (GLFF) to fuse multi-scale global features from the whole image and subtle local features from multiple informative patches. An Attention-based Multi-scale Feature Fusion (AMSFF) module is developed for feature fusion and a novel Patch Selection Module~(PSM) is designed for local patch selection.
    
    \item We construct a large-scale and highly-diverse  {AI-synthesized face image dataset} $DF^3$ by considering 6 state-of-the-art generation models and 5 post-processing operations to approach the real-world applications. 
    To our best knowledge, this is the first dataset considering both advanced generation models and diverse post-processings for challenging synthesized image detection. 
    
    
    \item 
    Extensive experimental results demonstrate the difficulty of detecting generated faces in the proposed $DF^3$ dataset for existing detection methods, with a performance degradation of over $30\%$. Our method achieves outstanding generalization ability on the $DF^3$ dataset and several public datasets. 
 
\end{itemize}

A preliminary version of this work was in~\cite{ju2022fusing}, in which we proposed a two-branch model to combine global information and subtle local features to detect diverse types of AI-synthesized images. This work extends on that in two ways: 1) In the global branch, inspired by recent work~\cite{zhao2021multi} revealing that slight artifacts caused by forgery methods tend to be preserved in the shallow features, we extract and fuse low-level features with high-level semantic features for more fine-grained feature learning. Based on the fused global feature, our local branch selects informative patches for the classification. 2) To evaluate the performance of the proposed method under real-world applications, we construct a challenging  {AI-synthesized face image dataset} $DF^3$ as a benchmark by considering 6 state-of-the-art generation models and 5 post-processing operations to simulate the real-world applications. 

The remainder of the article is organized as follows. Related works are briefly reviewed in Section~\ref{sec:relatedwork}. The proposed GLFF method for {synthesized image detection} is presented in Section~\ref{sec:method}. The $DF^3$ dataset is introduced in Section~\ref{sec:dataset}. Section~\ref{sec:exp} shows the experimental results and corresponding analysis. The conclusion and future work are presented in Section~\ref{sec:conclusion}.


\vspace{-0.4cm}
\section{Related Work}
\label{sec:relatedwork} 
 {This section gives a brief overview of the research literature on deep learning-based image synthesis and detection techniques. It is worth noting that forensic research on detecting fake media has primarily focused on two tasks. The first task centers around detecting synthetic images that have been entirely generated by AI models at the pixel level. The second task involves detecting DeepFake videos where identities have been replaced. Existing studies have tended to investigate these two tasks separately due to their significant differences in nature. These differences include the forgery region, which can either be the entire image or just the face; the type of generation model used, such as GANs or Diffusion models for synthetic images and AutoEncoders for DeepFake videos; and the media modality, which can be a single image or a video sequence. As far as we know, there is currently no available model that can effectively handle both tasks with promising performance. Our work focuses on synthetic image detection due to the rapid growth and high diversity of image generation techniques in this field.}

\vspace{-0.3cm}

\subsection{Image Generation}

 {The development of deep learning techniques, especially the Generative Adversarial Networks (GANs)~\cite{goodfellow2020generative}, Variational Autoencoders (VAEs)~\cite{kingma2013auto}, and Diffusion models~\cite{rombach2022high}, has made image generation increasingly realistic. GANs employ a training strategy that entails a min-max competition between a generator network and a discriminator network~\cite{goodfellow2020generative}. A lot of GAN models, such as ProGAN~\cite{karras2017progressive}, StyleGAN 1/2/3~\cite{karras2019style,karras2020analyzing,karras2021alias}, CycleGAN~\cite{zhu2017unpaired}, Pix2Pix~\cite{isola2017image}, BigGAN~\cite{brock2018large}, StyleGAN-T~\cite{sauer2023stylegan}, \etc, have been introduced for generating images based on conditions such as class label, text, or no conditioning. Besides GAN models, VAE~\cite{kingma2013auto,van2017neural} and its variants are also utilized in Taming Transformers~\cite{esser2021taming} and DALL·E~\cite{ramesh2021zero} for realistic image generation. Recently, Diffusion models~\cite{ho2020denoising,karras2022elucidating,vahdat2021score,dhariwal2021diffusion,rombach2022high,nichol2021glide,ramesh2022hierarchical} have gained popularity and achieved state-of-the-art results in image synthesis, due to their favorable features such as stable training and diverse modes compared with previous models.} 

\vspace{-0.3cm}
 {\subsection{AI-synthesized Image Detection}}

 {A variety of detection methods have been developed to distinguish between real and synthetic images. A common approach is to train deep neural networks to extract various features to solve the binary classification task~\cite{wang2020cnn,gragnaniello2021gan,asnani2021reverse,zhao2021multi,guo2021robust}. We categorize them into three groups according to the features they extract for classification.} 

 \noindent{\textbf{Spatial Feature Learning.}}
 {Several studies have developed detectors that differentiate synthetic images from real ones by extracting spatial features from RGB inputs~\cite{wang2020cnn,marra2018detection,chai2020makes,schwarcz2021finding,yu2019attributing}. Some of these approaches only extract global features from the entire image, ignoring local artifacts~\cite{wang2020cnn,marra2018detection,yu2019attributing}. To obtain more generalized features for diverse manipulation detection, recent works demonstrated the importance of low-level features and features from local patches for detection and localization~\cite{zhao2021multi,chai2020makes,schwarcz2021finding,zhang2021thinking,gragnaniello2021gan,yu2022patch,ju2022fusing,mandelli2022detecting}. Specifically, Zhao \etal~\cite{zhao2021multi} observe that the slight artifacts caused by forgery methods tend to be preserved in the high-frequency component of the shallow features. Therefore, they propose a multi-attentional framework to solve deepfake detection by incorporating shallow features. The study in \cite{chai2020makes} proposes a patch-based classifier and obtains good generalization ability. The method in \cite{schwarcz2021finding} also extracts features from various facial parts. However, this method utilizes local artifacts without considering the global semantic features, which may not generalize well to more diverse and subtle synthesized image detection tasks. }

 \noindent{\textbf{Frequency Feature Learning.}}
 {Artifacts present in various generative models have been identified through frequency-domain analysis, which can serve as a means to differentiate between real and generated images~\cite{zhang2019detecting,frank2020leveraging,durall2020upconv,gragnaniello2021gan,corvi2022detection,corvi2023intriguing,qian2020thinking,durall2019unmasking}. Representatives such as spectrum magnitude, 2d-FFT, and 2d-DCT serve as input features for CNN binary classification~\cite{durall2019unmasking,zhang2019detecting,frank2020leveraging,qian2020thinking}. However, the filters employed in these works are typically predetermined, making them inadequate for dynamically detecting fake images from unseen models. Additionally, frequency domain analysis can significantly deteriorate when testing images are subjected to post-processing techniques.}

 \noindent{\textbf{Feature Fusion.}}
 {Numerous works, which rely on feature fusion, have been proposed for computer vision tasks, including semantic segmentation~\cite{fan2021rethinking,zhang2018exfuse,minaee2021image} and object detection~\cite{wei2020f3net,zou2023object,liu2020deep,zhang2020weakly,zhuge2022salient}. Different from them, our model is designed for AI-synthesized image detection. In the field of fake image detection, several methods are proposed for extracting and fusing various complementary features. Specifically, dual-color (RGB and YCbCr) feature fusion~\cite{chen2021robust} is utilized for robust generated image detection. Frequency features from global and local regions are also fused and prove the effectiveness of frequency analysis in detecting fake images~\cite{qian2020thinking}. Spatial feature fusion methods are also proposed to fuse artifacts from multiple local patches and features from global images for synthesized image detection~\cite{zhang2021thinking, chen2022distinguishing,mandelli2022detecting}. However, their local patches are pre-defined by dividing the image evenly or by detecting the landmarks but our method introduces PSM to select the patches automatically.}

The motivation for fusing global and local features for AI-synthesized image detection is that AI-synthesized images exhibit artifacts in both global spatial patterns (such as texture inconsistency and color distribution~\cite{wang2022gan}) and local informative patches (like boundary artifacts and misplaced objects~\cite{yu2022improving}). To enhance the detector's generalization ability to the synthesized images, particularly those produced by advanced generation models or processed with diverse post-processing operations, we propose to extract discriminative representations by combining multi-scale global features from the whole image with refined local features from informative patches. Although global-local fusion has been widely used in vision tasks, different from existing works, our global and local fusion method fuses multi-scale features at two distinct levels: the model level (\ie, hierarchical features extracted from shallow and deep layers) and the image level (\ie, features extracted from the entire image and automatically cropped local patches). This enables us to extract subtle and discriminative signal artifacts in an adaptive manner, which facilitates the detection of generalized AI-synthesized image detection.



\begin{figure*}[tb] 
\centering 
\includegraphics[width=0.95\textwidth]{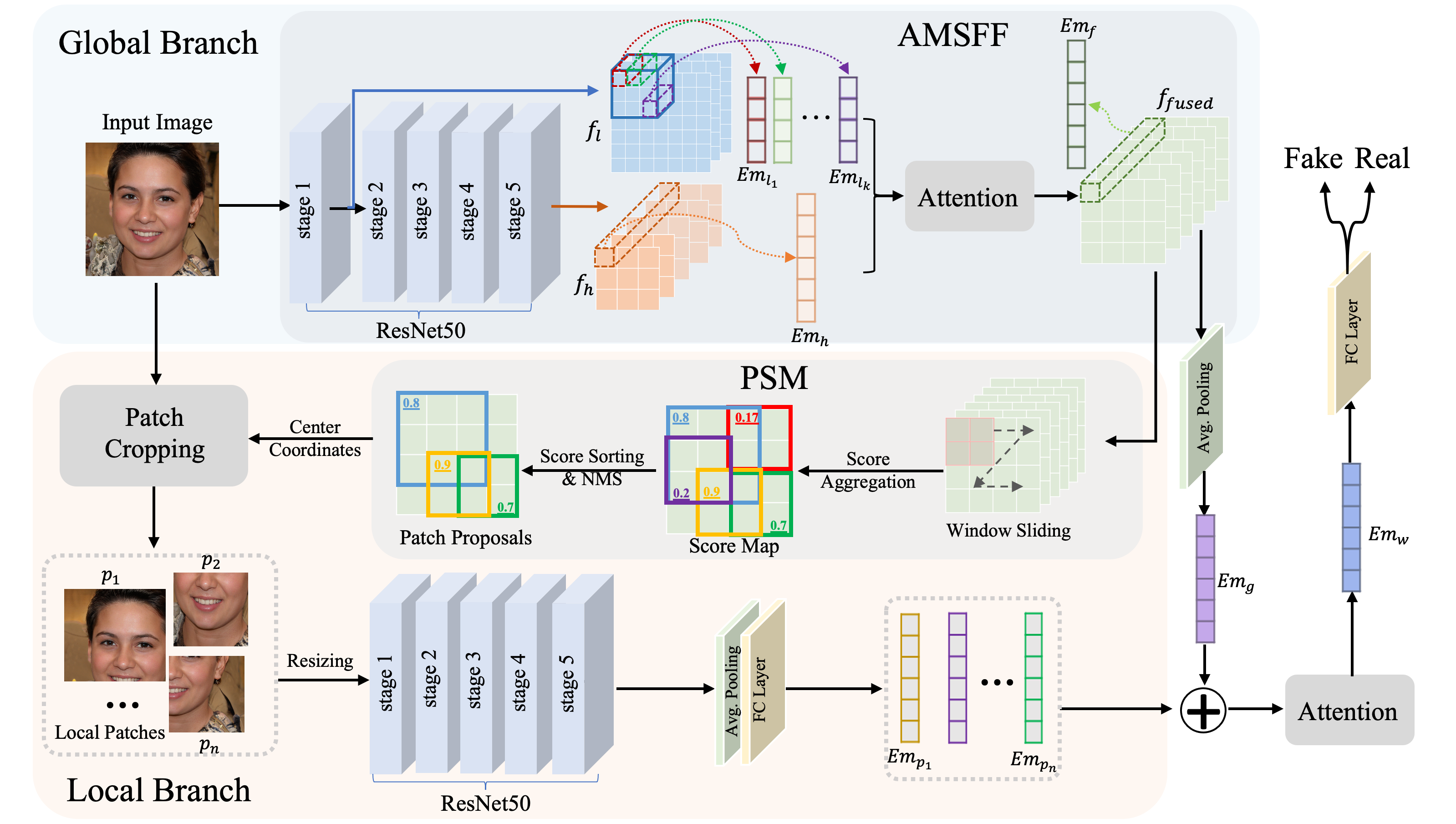}
\vspace{-0.2cm}
\caption{\textbf{The architecture of GLFF.} The global branch uses \textbf{A}ttention-based \textbf{M}ulti-\textbf{s}cale \textbf{F}eature \textbf{F}usion module~(AMSFF) to extract low-level and high-level spatial features, and the local branch extracts subtle feature from local patches selected by \textbf{P}atch \textbf{S}election \textbf{M}odule~(PSM). The attention mechanism is utilized to fuse global and local embeddings for classification.} 
\vspace{-0.4cm}
\label{fig:mainfig} 
\end{figure*}

\vspace{-0.35cm}
\subsection{AI-synthesized Image Datasets}
There have been several publicly available datasets of synthesized images. 
Karras \etal~\cite{karras2019style} released 100k-Generated-Images including 100k synthetic face images generated using their proposed StyleGAN~\cite{karras2019style} trained on FFHQ dataset~\cite{karras2019style}. 100k-Faces~\cite{100k} is another dataset containing 100k synthetic images generated using StyleGAN~\cite{karras2019style} but trained on a more diverse dataset. Dang \etal~introduced the Diverse Fake Face Dataset~(DFFD) generated using StyleGAN~\cite{karras2019style} and ProGAN~\cite{karras2017progressive} models in \cite{dang2020detection}. Neves \etal~proposed the iFakeFaceDB database~\cite{neves2020ganprintr} using StyleGAN~\cite{karras2019style} and ProGAN~\cite{karras2017progressive}. Different from previous datasets, the iFakeFaceDB dataset is processed with an anti-forensics algorithm GANprintR~(GAN Fingerprint Removal) to remove the artifacts and thus confuse fake detectors. Want \etal~\cite{wang2020cnn} collected a diverse dataset created from 11 synthesis models, including GANs, models with perceptual loss, low-level vision models, and DeepFakes to evaluate the generalization ability of their proposed detection method. Recently, Asnani \etal~presented a fake image dataset with 116K images generated by 116 different generative models for deepfake detection and model attribution~\cite{asnani2021reverse}. 

Existing datasets adopted generative models as the only tool to generate fake images without considering any post-processing operations in real-world applications. Although the OpenForensics dataset~\cite{le2021openforensics} utilizes several post-processing operations to blend GAN-generated faces into real images, and iFakeFaceDB dataset~\cite{neves2020ganprintr} considers the anti-forensics application, they only consider limited post-processing operations, which is far different from real-world scenarios. Moreover, state-of-the-art models such as transformer-based or diffusion-based models are not covered in existing datasets. Therefore,
impressive performance on these datasets can not ensure that current detection models can work successfully under a more challenging scenario where the manipulated images are generated not only by advanced models but also with well-designed post-processings. 

\vspace{-0.35cm}

\section{GLFF for  {Synthesized Image Detection}}
\label{sec:method}

To learn generalized representations for synthesized image detection, we emphasize the importance of learning rich and complementary features in detecting diverse face forgeries and propose the Global and Local Feature Fusion (GLFF) framework by combining multi-scale global representations with informative local features. In this section, we first introduce the overall architecture of our detection method, and then elaborate on each module of the method, including the global branch with a novel feature fusion scheme, the local branch with a patch selection module, and the final classification module.  

\begin{figure}[!t] 
\centering 
\includegraphics[width=0.48\textwidth]{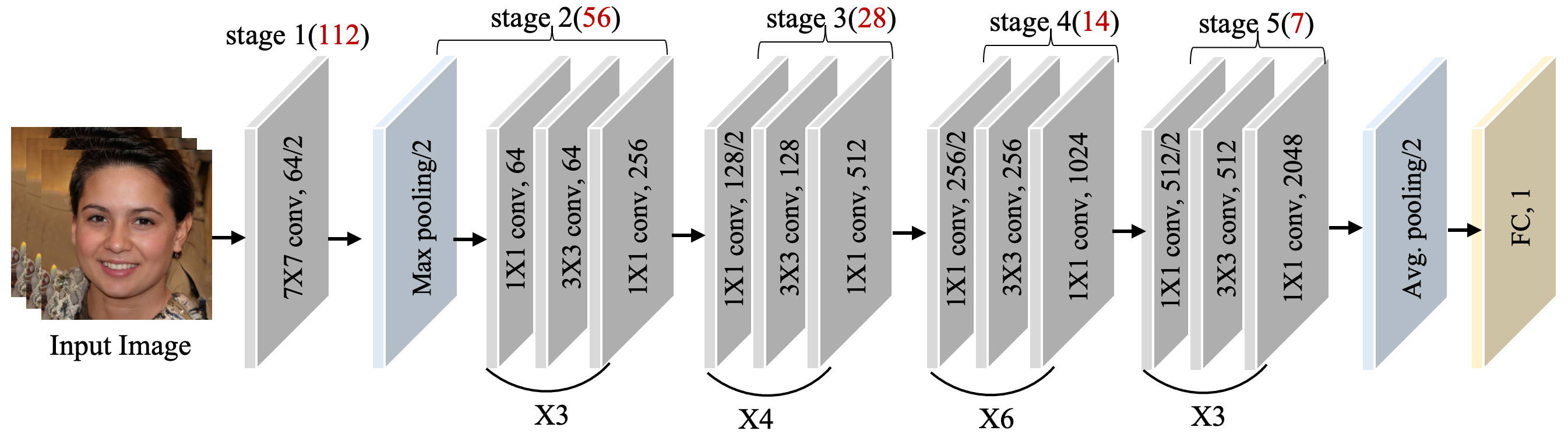}
\vspace{-0.2cm}
\caption{\textbf{The architecture of ResNet-50 in our AMSFF module.}} 
\vspace{-0.4cm}
\label{fig:resnet50} 
\end{figure}

\vspace{-0.4cm}
\subsection{Overall Architecture}

The overall structure of our GLFF model is illustrated in Fig.~\ref{fig:mainfig}. In the global branch, the input image is fed into the backbone network, which is based on the ResNet-50~\cite{he2016deep} in our implementation, considering its concise architecture and competitive performance in various face-related classification tasks~\cite{wang2020cnn, jia20203d, li2018exposing}. The Attention-based Multi-scale Feature Fusion~(AMSFF) module is designed to aggregate low-level and high-level features extracted from the ResNet-50 model for global representation. To extract informative local features, the fused global representation is then fed into the local branch, from which the proposed Patch Selection Module~(PSM) first selects informative patches with higher energy, and thus higher contribution to the classification~\cite{zhang2021multi}. The local features are extracted from these selected patches and then fused with the global representation with an attention-based module for the final binary classification. 

\vspace{-0.4cm}

\subsection{Global Branch}
The global branch in our model is designed to learn multi-scale global features and guide local branch patch selection. Inspired by the previous findings that enhancing the feature from shallow layers can prevent the subtle difference from disappearing in the deep layers for face-swapped deepfake detection~\cite{zhao2021multi}, we develop an attention module to aggregate hierarchical features to fully explore the subtle artifacts. Different from previous works which combine different features based on concatenation~\cite{liu2019calf} or bilinear pooling~\cite{zhao2021multi}, we fuse multi-scale features by designing an Attention-based Multi-scale Feature Fusion~(AMSFF) module based on the multi-head attention mechanism~\cite{vaswani2017attention}. 

Given an input image, the AMSFF module employs the ResNet50~\cite{he2016deep} to extract low-level feature maps $F_l$ from the shallow convolutional layer~(stage 1 in Fig.~\ref{fig:mainfig}) and high-level semantic feature maps $F_h$ from the deep convolutional layer~(stage 5 in Fig.~\ref{fig:mainfig}). The stage represents a group of convolutional and pooling layers in ResNet-50 and more details about the layers in each stage are shown in Fig.~\ref{fig:resnet50}. Assuming the size of $F_h$ be $H_h\times W_h \times C_h$, for each pixel in $F_h$, the AMSFF aggregates the values along the channel dimension as a vector $E_{m_h}$, whose size is $1\times 1 \times C_h$. Assuming that the size of $F_l$ is $H_l \times W_l \times C_l$, we divide $F_l$ into patches evenly without overlap, and the size of each patch is $\frac{H_l}{H_h} \times \frac{W_l}{W_h}$. In the cropped patch, each pixel vector is taken as an embedding $E_{m_{l_i}} (i \in [1, k], k$ is the number of pixels in the cropped patch). 
To effectively fuse the low-level and high-level features, AMSFF concatenates the $k$ embeddings from low-level feature maps and $1$ embedding from high-level feature maps to learn the correlations among these embeddings and combine them as a fused embedding $E_{m_f}$ with the learned weights. Specifically, the concatenated $k+1$ embeddings are inputted into AMSFF as the query $\mathrm{Q}$, the key $ \mathrm{K}$, and the value $\mathrm{V}$. Multi-head attention with $s$ heads is used. The $i$th attention head $\mathrm{h}_i (i\in [1,s])$ is computed with the scaled dot product attention as following~\cite{vaswani2017attention}:
\begin{equation}
\vspace{-0.2cm}
\small
\label{eqn:attention}
\mathrm{h}_{i}=softmax(\frac{{\mathrm{W}_i}^Q\mathrm{Q}\times{{(\mathrm{W}_i}^K\mathrm{K})}^T}{\sqrt{d}})\times {\mathrm{W}_i}^V\mathrm{V}
\end{equation}
where ${\mathrm{W}_i}^Q$, ${\mathrm{W}_i}^K$, ${\mathrm{W}_i}^V$ are parameter matrices and $d$ is the dimension of $K$. The attention scores of all heads are concatenated as the input of a linear transformation to obtain the value of multi-head attention as follows~\cite{vaswani2017attention}:

\vspace{-0.2cm}
\begin{equation}
\small
\label{eqn:mha}
E_{m_f} = Concat(\mathrm{h}_i)\mathrm{W}^o (i \in [1,n])
\end{equation}
where $\mathrm{W}^o$ is also a parameter matrix. The fused one-dimensional vector $E_{m_f}$ is taken as a pixel vector on the fused feature maps. After iterating from all the pixels on the high-level feature maps, the fused feature maps, denoted as $F_{fused} \in \mathbb{R}^{H_h \times W_h \times C_h}$ with $C_h$ channels and $H_h\times W_h$ spatial size. The shape of the fused feature maps is the same as the high-level feature maps, but with more dense information that each pixel contains fused information from a high-level embedding and several low-level embeddings. The global embedding $E_{m_g}$ is finally acquired by feeding $F_{fused}$ into an average pooling layer.

\vspace{-0.4cm}
\subsection{Local Branch with Patch Selection Module}
Our local branch is designed to select the most $n$ informative patches (see $p_1, p_2, ..., p_n$, in Fig.~\ref{fig:mainfig}) for extracting discriminative local embeddings $E_{m_{p_1}}, E_{m_{p_2}}, ..., E_{m_{p_n}}$. Assuming that regions with higher energy in the global fused feature maps $F_{fused}$ typically provide more information for classification tasks~\cite{zhang2021multi}, we introduce the Patch Selection Module~(PSM) to select informative patches by calculating the patch score from the fused feature maps $F_{fused}$. 

During the calculation of the patch score, the activation map $A$ is estimated by aggregating $F_{fused}$ along the channel dimension as shown as $A(x,y) = \sum_{j=1}^{C_h} F_{fused}^j(x,y)$.
Then the average score of a window on the activation map indicates its ``informativeness'' on the input image. Specifically, a sliding window with $H_{w} \times W_{w}$ spatial size is used to aggregate the information on the activation map $A$, and the score of each window $S_{w}$ is calculated as follows:

\vspace{-0.2cm}
\begin{equation}
\small
\label{eqn:sw}
S_{w} = \frac{\sum_{x=0}^{W_{w}-1}\sum_{y=0}^{H_{w}-1}A(x, y)}{W_{w} \times H_{w}}
\end{equation}

By sorting the scores from all windows and reducing patch redundancy with non-maximum suppression~(NMS), $n$ windows are selected as the patch proposals. The central coordinates of these patch proposals are mapped back to the input image to guide the patch cropping. After patch cropping, the most $n$ informative local patches, $p_1, p_2, ..., p_n$, are fed into the ResNet50 for local embeddings $E_{m_{p_1}}, E_{m_{p_2}}, ..., E_{m_{p_n}}$ extraction.

\vspace{-0.4cm}

\subsection{Feature Fusion for Classification}
The extracted global embedding $E_{m_g}$ from the global branch and local embeddings $E_{m_{p_1}}, E_{m_{p_2}}, ..., E_{m_{p_n}}$ from informative $n$ local patches in the local branch are concatenated and fed into the multi-head attention module for fusion. Following the Equations (\ref{eqn:attention}) and (\ref{eqn:mha}), the final embedding $E_{m_w}$ is calculated for the final binary classification. We used the Softmax classifier and Binary Cross Entropy loss function in our implementation. 

\vspace{-0.2cm}
\section{$DF^3$ Dataset}
\label{sec:dataset}

Existing  {AI-synthesized image datasets contain images generated with limited generation models without considering well-defined post-processings. To construct a highly diverse dataset that approaches real-world applications, we create the $DF^3$ dataset with newly emerged generation models and a variety of post-processing techniques. We only consider face image generation in the $DF^3$ dataset for the following two reasons. Firstly, face images often contain sensitive information, making them more vulnerable to being exploited for malicious purposes. Secondly, it is easier to construct and manipulate a dataset with face objects since there are numerous models available with powerful generation capabilities.} In this section, we will first introduce basic statistical information about the dataset, and then give details of the dataset construction.

\begin{table*}
\renewcommand\arraystretch{1}
\center
\footnotesize
\newcommand{\tabincell}[2]{\begin{tabular}{@{}#1@{}}#2\end{tabular}} 
\vspace{-0.4cm}
\caption{\label{tb:$DF^3$} \textbf{Statistical details of the proposed $DF^3$ dataset.}} 
\vspace{-0.4cm}
\scalebox{0.97}{
\begin{tabular}{ c | c c  c  c  c  c }
 \hline
    \textbf{Generation Model} & \textbf{\tabincell{c}{Protocol-1\\Unprocessed}}&  \textbf{\tabincell{c}{Protocol-2\\Common Post-processing}}& \textbf{\tabincell{c}{Protocol-3\\Face Blending}} &  \textbf{\tabincell{c}{Protocol-4\\Anti-forensics}} & \textbf{\tabincell{c}{Protocol-5\\Multi-image Compression}} & \textbf{\tabincell{c}{Protocol-6\\Mixed}}\\

\hline
  StyleGAN2~\cite{karras2020analyzing}, 2020 &   $1,000$  & $1,000$ & $994$ & $996$ & $1,000$ &  $2,652$\\
  \hline
  StyleGAN3~\cite{karras2021alias}, 2021   &  $1,000$  & $1,000$ & $995$ & $996$  & $1,000$ &  $2,609$\\
  \hline
  3DGAN~\cite{chan2022efficient}, 2022  &  $1,000$ &  $1,000$ & $984$ & $996$ & $1,000$ &  $2,588$\\
  \hline
  Transformers~\cite{esser2021taming}, 2021   & $961$ & $961$ & $959$ & $960$ & $961$ & $2,710$\\
  \hline
  LSGM~\cite{vahdat2021score}, 2021   &  $1,199$ & $1,199$ & $998$ & $1,191$ & $1,199$ &  $2,680$\\
  \hline
  Latent-diffusion~\cite{rombach2022high}, 2022 &  $1,000$ & $1,000$ &$998$ & $996$ & $1,000$ & $2,694$\\
  \hline
  Total & $6,160$ &  $6,160$&$5,928$ & $6,135$& $6,160$&  $15,933$\\
 \hline
\end{tabular}
}
\vspace{-0.4cm}
\end{table*}

\begin{figure}[tb] 
\centering 
\includegraphics[width=0.5\textwidth]{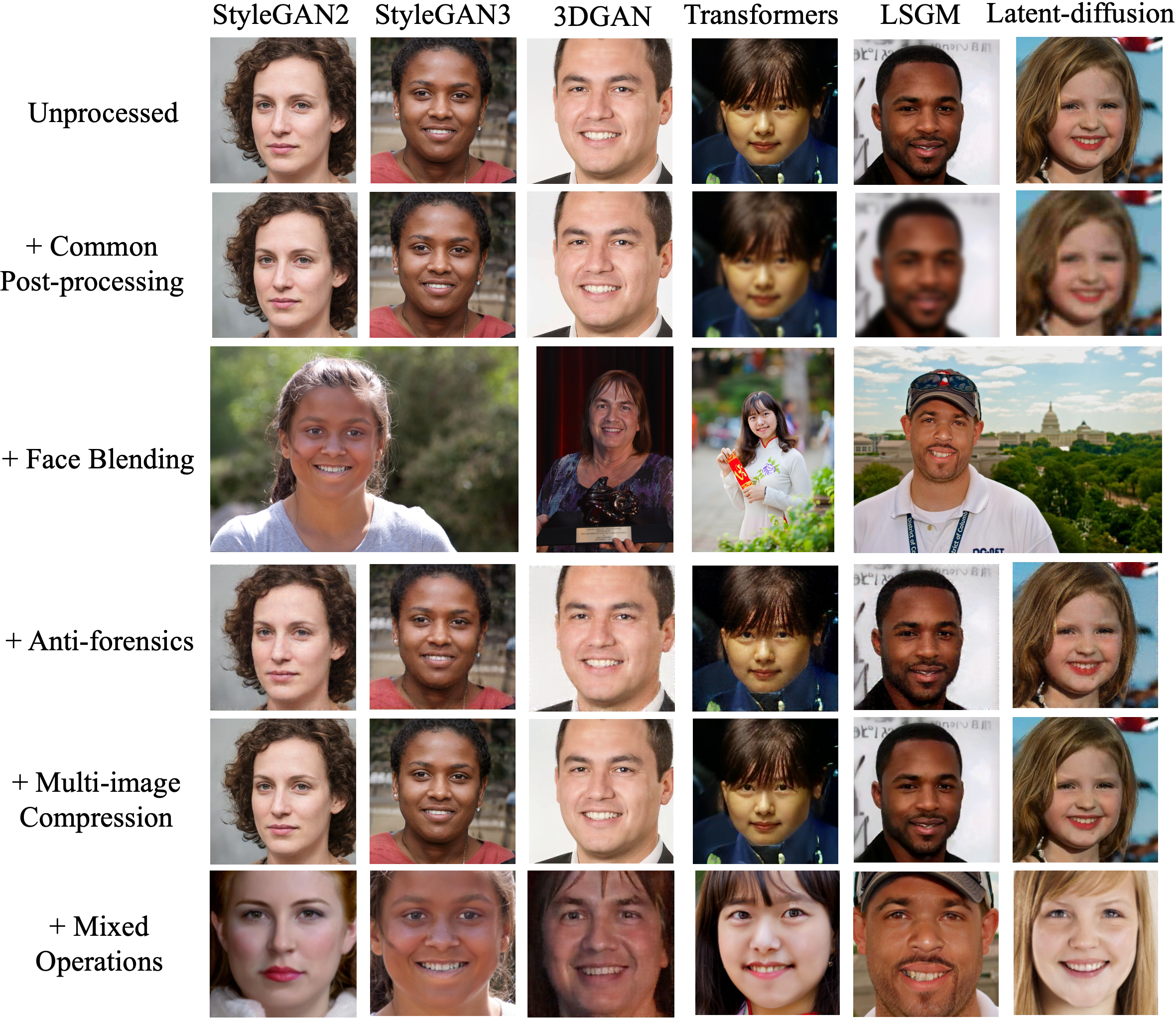}
\vspace{-0.4cm}
\caption{\textbf{ {Examples from the $DF^3$ dataset (best viewed online in color with zoom-in).  From up to down, the original image is generated by a generation model followed by five post-processing operations.}}} 
\vspace{-0.4cm}

\label{fig:ourdataset} 
\end{figure}

The $DF^3$ dataset contains approximately 46,400 fake face images from 6 generation models and 5 post-processing protocols. The number of images in each protocol and model category are shown in Table~\ref{tb:$DF^3$}. Examples of the generated faces can be found in Fig.~\ref{fig:ourdataset}.

Our dataset construction consists of two steps: 1) Unprocessed fake image generation; 2) Post-processing implementation. For unprocessed fake image generation, we selected $6$ state-of-the-art generative models, \ie, $3$ unconditional GANs: StyleGAN2~\cite{karras2020analyzing}, StyleGAN3~\cite{karras2021alias}, 3DGAN~\cite{chan2022efficient}, $1$ Transformers~\cite{esser2021taming}, and $2$ diffusion models: LSGM~\cite{vahdat2021score} and Latent-diffusion~\cite{rombach2022high}. We use the following two criteria in selecting these models: 1) they show impressive performance on face generation; 2) they all provide open-source pre-trained models that are trained on large-scale and high-quality face images. Specifically, we generated around 1,000 face images with each model. 
For the post-processing implementation, different from existing studies which merely consider image compression or blurring that can be easily addressed by data augmentation during model training, we include more diverse post-processing operations to create challenging face forgeries as well as to promote more robust detection methods.
Considering that the artifacts in generated face images can be suppressed or destroyed by both common post-processing and well-designed anti-forensics, we propose 5 protocols, including common post-processing (\eg, JPEG Compression and Gaussian Blur caused by image laundering), face blending that merges a fake face to a real background to improve authenticity, anti-forensics with well-designed adversarial attacks, multi-image compression to add complex noises, and mixed version of the previous operations. 

\subsubsection{Common Post-processings}
Two popular post-processing methods are included for this protocol: JPEG Compression and Gaussian Blur considering that they are widely used in image laundering by social media platforms to minimize the bandwidth and accelerate the transmission. 
Given a generated face image, we randomly select one set from three common post-processing operations, \ie, JEPG Compression, Gaussian Blur, and Gaussian Blur combined with JPEG Compression. In JPEG Compression, the quality parameter is discretely selected randomly from $[20, 90]$. In Gaussian Blur, the input image is blurred with parameter $\sigma$ continuously selected from $[1, 4]$ randomly. Gaussian Blur combined with JPEG Compression conducts Gaussian Blur followed by JPEG Compression with the same parameter selection strategies.

\subsubsection{Face Blending}
Current fake image generation models usually generate realistic faces with a large face region but without any semantically meaningful background (as shown in the first row of Fig.~\ref{fig:ourdataset}), which limits the application of fake faces and also becomes an important clue for detection. 
To improve the diversity and the authenticity of fake faces, we propose to blend the fake faces into various real-world scenes to generate more complex and living photos. In the face blending subset, we use the SimSwap~\cite{DBLP:conf/mm/ChenCNG20} to swap our fake faces from 6 generation models to the real-world face pictures in FFHQ-unaligned dataset~\cite{karras2019style}, which has a high diversity in face region and semantic background. In total, 5,928 images are generated under this protocol after cleaning the failed samples manually, and more examples can be found in Fig.~\ref{fig:ourdataset}. To the best of our knowledge, this is the first dataset to combine fake faces with real-world scenes. We believe this kind of manipulation is significant while challenging to detect due to the following two reasons: 1) This manipulation enhances the realism of synthetic faces and extends the range of applications for artificial faces (not only used as account profiles); 2) Such face-swap manipulation improves the difficulty in detection due to the diverse face regions and complex post-processing from color tuning and blending steps which may weaken the artifacts of the original generated images and introduce new noises. 


\subsubsection{Anti-forensics}
The design of effective adversarial attacks is another widely studied method to challenge the detection of fake faces. To evaluate how detection methods can perform under this well-designed post-processing, we include two popular algorithms as the anti-forensics method to confuse current detectors, including the CW attack algorithm~\cite{carlini2017towards} for adversarial examples generation and the GANPrintR~\cite{neves2020ganprintr} for GAN image artifacts removing.
Specifically, we adopt the CW algorithm to attack a basic CNN-based detection model~\cite{wang2020cnn}. As there are many detection models that are developed from this basic CNN model, we attack this basic CNN detection model to explore if this attack works for other models with similar architecture. Another anti-forensics method GANPrintR~\cite{neves2020ganprintr} utilizes an auto-encoder as a low-pass filter to remove more high-frequency components in the GAN-generated images, which has been proven quite effective in misleading the detector to make wrong decisions. We randomly apply either the CW attack algorithm or GANPrintR to the generated face image and generate a total of 6,135 images for this protocol. Fig.~\ref{fig:ourdataset} shows the high quality of images in this subset.

\subsubsection{Multi-image Compression}
Considering that single-image compression can be easily addressed by data augmentation in the training process, we propose a novel multi-image compression strategy inspired by the H.164 video compression, so that we can not only compress the single image in the spatial domain but also compress temporal redundancy of similar blocks between frames sequentially. High-frequency information has been proven to be an important cue for GAN-generated image detection and attribution~\cite{asnani2021reverse,zhang2019detecting,frank2020leveraging}. We believe that the combination of intra-image compression with cross-image compression will further suppress the high-frequency information in fake images due to the reduction of high redundancy in both spatial and temporal domains.
Specifically, we randomly group multiple synthesized images as a video sequence and compress it with H.264 using HandBrake~\cite{Handbrake} tool with the default quality parameter as $22$. Then the processed images are obtained by frame extraction to compose the subset with 6,160 images. 

\subsubsection{Mix Subset}
We finally mix the previous 4 types of post-processing operations to further approach real-world applications where the fake images can experience multiple manipulations to significantly fool the detection tool. 
Given a generated image, we randomly select several manipulations out of the $4$ manipulations and consider the following combinations: Face Blending~+~Common Post-processing, Face Blending~+~Multi-image Compression, Multi-image Compression~+~Anti-forensics, Face Blending~+~Common Post-processing+Anti-forensics and Face Blending~+~Multi-image Compression~+~Anti-forensics. A total of 15,933 images are generated for this protocol. 


\vspace{-0.2cm}
\section{Experiments}
\label{sec:exp}
In this section, we conduct extensive experiments to evaluate the effectiveness of the proposed detection method. We first introduce the implementation details of our method, followed by state-of-the-art baselines for comparison and several evaluation datasets. Then we compare and analyze the experimental results of our method on four datasets. Finally, we present the ablation studies and visualization results.

\vspace{-0.4cm}

\subsection{Implementation Details}
All experiments are conducted on 4 NVIDIA RTX A5000 GPU cards. Our model is implemented with Pytorch~\cite{paszke2019pytorch} and trained with a batch size of $64$ and base learning rate of $0.0001$. The Adam optimizer and the Binary Cross Entropy loss function are used. ResNet-50~\cite{he2016deep}, pre-trained with ImageNet, is utilized as the backbone network to extract features from different scales. We utilize the same ResNet50 model in both global and local branches to ensure time efficiency. The input image is resized to $224\times224 \times 3$. Before resizing, the input image is blurred with $\sigma$ $\sim$ Uniform $[0, 3]$ and JPEG-ed with quality $\sim$ Uniform $\{30, 31, . . . , 100\}$ with $10$\% probability during training process. Fused feature maps $F_{fused}$ with $2048$ channels and $7\times 7$ spatial size are inputted into the local branch to guide PSM. In PSM implementation, two sliding windows with the spatial size of $3 \times 3$ and $2 \times 2$ are used to aggregate the score of each local patch on the activation map. $6$ local patches are selected including $3$ patches from window size $3 \times 3$ and $3$ patches from window sizes $2 \times 2$. To utilize patches with different resolutions, we map the patches from window size $3 \times 3$ into $224 \times 224$ while the patches from window sizes $2 \times 2$ are mapped into $112 \times 112$ on the input image. These $6$ patches are further resized into $224 \times 224$ and fed into ResNet-50 for local embedding extraction. The dimension of extracted global embedding and local embedding is set as $128$. In the AMSFF module and the attention-based global and local feature fusion module, $3$ attention layers with $4$ heads are utilized to combine all the extracted features. 

\vspace{-0.4cm}

\subsection{Baseline Methods}
We use $5$ state-of-the-art generated image detection models as the baselines, of which $3$ models are based on spatial information: CNN-aug~\cite{wang2020cnn}, Nodown~\cite{gragnaniello2021gan}, and PSM~\cite{ju2022fusing} and $2$ are based on spectral information: GAN-DCT~\cite{frank2020leveraging} and BeyongtheSpectrum~\cite{yang_ijcai21}. We followed the settings in their paper and trained them on our training dataset from scratch except for Nodown, in which the training code was not provided so we used their released pre-trained weights.

For evaluation metrics, we utilize commonly used metrics including Overall Accuracy~(OA), which represents the probability of an image being accurately classified by a detector, and Area Under ROC Curve~(AUC), which calculates the area under the ROC curve, a graph showcasing the true positive rate (TPR) versus the false positive rate (FPR) across different threshold settings, to evaluate our model and the baselines. We also report the Average AUC~(Avg. AUC) by averaging it over the 6 protocols in our experiment. Moreover, ROC Curves are shown for a qualitative comparison. 

\vspace{-0.4cm}

\subsection{Datasets}
The training dataset we used for our model training is the same as that in~\cite{wang2020cnn}. It comprises 362K real images from the LSUN dataset~\cite{yu2015lsun} with $20$ object classes, including person, cat, dining room, \etc,  {The LSUN dataset was chosen for two main reasons. Firstly, fake images often display similar artifacts caused by the image generation process, which are not connected to the images' semantic content. Secondly, Wang \etal~(2020)~\cite{wang2020cnn} found that increasing the diversity of training images can improve generalization performance}. 362K images generated with ProGAN~\cite{karras2017progressive} are provided as fake images. The spatial size of all training images is $256 \times 256 $. 

 {The testing dataset consists of three parts. The first part is from the proposed challenging $DF^3$ dataset. This dataset is divided into $6$ subsets according to different post-manipulations. More details can be found in Table~\ref{tb:$DF^3$}. The second part comes from existing datasets (\ie, CNN-aug~\cite{wang2020cnn} and ReGMs~\cite{ asnani2021reverse}) considering that they contain diverse model-generated images. The third part is that we collect a small dataset of synthetic images generated by $5$ newly released generation models, including StyleGAN2-distillation~\cite{viazovetskyi2020stylegan2}, StyleGAN3~\cite{karras2021alias} and the state-of-the-art diffusion models such as DALL·E 2~\cite{ramesh2022hierarchical}, Guided-diffusion~\cite{dhariwal2021diffusion}, and also including partially manipulated images using background matting method~\cite{MODNet} to blend the generated images to a real background.} 

 \begin{table}[t]
\renewcommand\arraystretch{1}
\center
\vspace{-0.3cm}
\caption{\label{tb:existingtestingdataset} \textbf{Details of the highly-diverse testing datasets.}} 
\vspace{-0.3cm}

\scalebox{0.79}{
\begin{tabular}{c | c | c | c | c}
 \hline
 \textbf{Dataset Source} & \textbf{\# Models} & \textbf{\# Images(Real/Fake)} & \textbf{Object}\\
 \hline
  \hline
 \textbf{$DF^3$ (ours)} & 6 & 46,476 (0/46,476) & Face \\
 \hline
 \textbf{CNN-aug~\cite{wang2020cnn}} & 13 & 90,329 (45,169/45,160) & Face, Scenery, Animal, etc\\
\hline
 \textbf{ReGMs~\cite{asnani2021reverse}} & 13 & 37,709 (24,709/13,000) & Face, Scenery, Animal, etc\\
\hline
 \textbf{Newly-released models} & 5 & 12,388 (0/12,388) & Face, Scenery, Animal, etc\\
\hline
\end{tabular}}
\vspace{-0.4cm}
\end{table}

Since our $DF^3$ dataset only consists of fake images, we additionally collect 1,000 real images as negative samples to enable the metrics calculation such as AUC. These real images are randomly selected from the FFHQ dataset~\cite{karras2019style} for two reasons: 1) it consists of high-quality images with considerable variation in terms of age, ethnicity, and image background and 2) it is widely used in several generation models as a training dataset~\cite{karras2019style}. During testing, these real images are taken as a negative class, and images generated by each generation model are taken as the positive class to calculate classification metrics. More details of the testing dataset in the second part are listed in Table~\ref{tb:existingtestingdataset}. It is worth noting that our experiments mainly focus on cross-domain testing to evaluate the generalization ability of different detection methods (as previous works do~\cite{wang2020cnn, gragnaniello2021gan}).

\vspace{-0.4cm}

\subsection{Evaluation on the $DF^3$ dataset}

\begin{table*}
\renewcommand\arraystretch{1}
\center
\footnotesize
\vspace{-0.4cm}

\caption{\label{tb:test2} \textbf{Evaluation results (AUC) on the proposed $DF^3$ dataset. The highest value is highlighted in black.}} 
\vspace{-0.2cm}

\scalebox{0.9}{
\begin{tabular}{ c | c | c | c | c | c | c | c}
 \hline
  \textbf{Test Data}  & \textbf{CNN-aug\cite{wang2020cnn}} &  \textbf{GAN-DCT~\cite{frank2020leveraging}} & \textbf{Nodown~\cite{gragnaniello2021gan}} & \textbf{BeyondtheSpectrum~\cite{yang_ijcai21}} & \textbf{PSM\cite{ju2022fusing}} & \textbf{GLFF(Ours)} \\
\hline
\hline
\textbf{Unprocessed} & 0.723 & 0.656 & \textbf{0.970} & 0.819 & 0.901 & 0.906 \\
  \hline
  \textbf{Common Post-processing} & 0.710 & 0.443 & 0.823 & 0.624 & 0.878 & \textbf{0.887} \\
  \hline
  \textbf{Face Blending} &  0.795 & 0.483 & 0.888 & 0.558 & 0.877 & \textbf{0.905} \\
 \hline
   \textbf{Anti-forensics} & 0.605 & 0.504 & \textbf{0.894} & 0.644 & 0.863 & 0.834 \\
 \hline
\textbf{Multi-image Compression} &  0.217 & \textbf{0.646} & 0.105 & 0.577 & 0.411 & 0.547 \\
 \hline
   \textbf{Mixed} &  0.528 & 0.497 & 0.468 & 0.470 & 0.724 & \textbf{0.801}\\
 \hline
  \textbf{Average} &  0.596 & 0.538 &0.691 & 0.616 & 0.775 & \textbf{0.813}\\
 \hline
\end{tabular}
}
\end{table*}

Table~\ref{tb:test2} reports the AUC scores on each subset of the proposed $DF^3$ dataset. To be noted, the AUC of each subset is the average AUC among 6 generation models. The average AUC among all 6 subsets is also listed. All the methods achieve high accuracy on the unprocessed image subset because these images are directly generated with the 6 generation models without any post-processing so the fingerprints can be spotted by the current detectors. Our method achieves an AUC of $0.906$, slightly worse than the Nodown~\cite{gragnaniello2021gan} method, which argues that avoiding down-sampling layers could save more artifacts to help the forgery detection. 

Compared with the performance on the unprocessed image, it can be seen that the performance of CNN-aug, PSM, and our model with data augmentation during training drops slightly while greatly on others on the common post-processing subset. This is because these models adopt JPEG Compression and Gaussian Blur to augment their data during model training. 

The performance of GAN-DCT, Nodown, and BeyondtheSpectrum drops greatly on the Face Blending subset. This is because the face blending process needs to adjust the color of the original generated image and thus remove some low-level generation artifacts, which makes it harder to detect. However, PSM and our model drop slightly on this subset because these two methods exploit details from local patches such that they are more robust to such partial manipulation. 

The performance of all methods on the Anti-forensics subset drops greatly compared with that of clean images because these anti-forensics algorithms add some noise or remove the artifacts from the original generated image and thus make it more challenging to detect. As the CW attack algorithm is utilized targeted at the CNN-aug detector, so we did not compare CNN-aug with other methods for a fair comparison. For other methods, our model is slightly worse than Nodown, which achieves the best performance on the Anti-forensics subset due to more artifacts saved by avoiding the down-sampling processing. 

As for the Multi-image Compression subset, we utilize video compression as a new compression strategy because of its high compression rate. It can be seen that this processing decreases the performance heavily of all methods except GAN-DCT. This compression method could find similar patches across a group of images and compress them together instead of compressing a single image such that more high-frequency artifacts in the generated image are removed. However, GAN-DCT exploits DCT to extract spectral features for detection, which may be less influenced by the compression method than other spatial-based methods. 

Finally, we test all the methods on the Mixed subset, which combines 2 or 3 above-mentioned processes randomly. It can be seen that all the methods achieve poor performance on this subset, which proves that our proposed Mixed subset is very challenging and can be utilized as a tool to evaluate the robustness of future detection methods. We also report the average AUC score of each method by averaging it across all the subsets. It can be seen that our model achieves the best performance in terms of Average AUC and AUC on most subsets. This demonstrates the better generalization ability and robustness of our model than other methods, which only use whole image information or only focus on features from small patches without considering multi-scale information combinations. 

\begin{figure*}
\centering 
\includegraphics[width=0.86\textwidth]{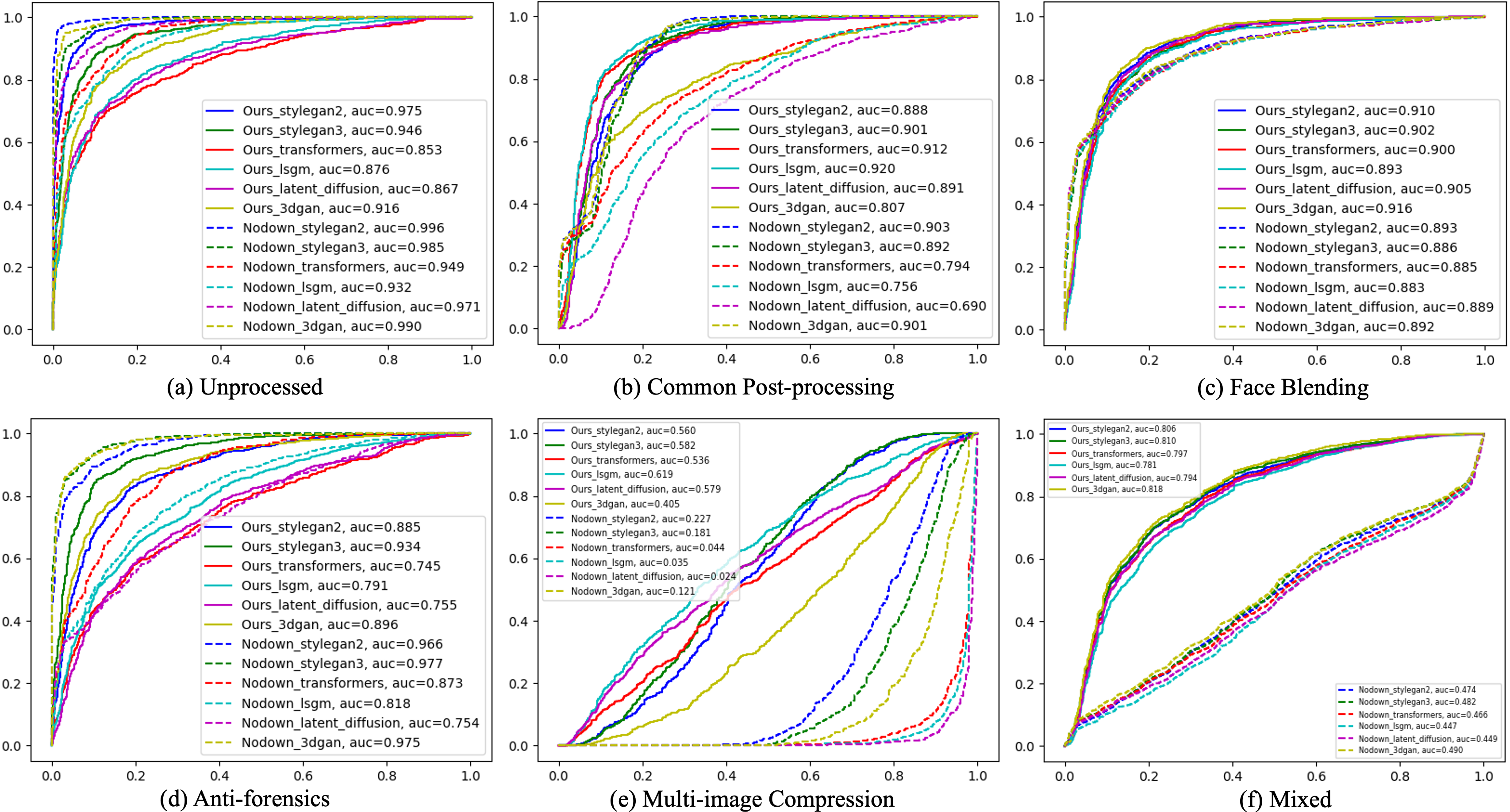}
\vspace{-0.3cm}

\caption{\textbf{Comparison of ROC curves under 6 generation models and 6 evaluation protocols in the $DF^3$ dataset.}}
\vspace{-0.4cm}

\label{fig:rocresults} 
\end{figure*}

To further demonstrate the superiority of our method qualitatively, we report the ROC curves on 6 generation models from 6 subsets in our $DF^3$ dataset. We compare our method with the Nodown model~\cite{gragnaniello2021gan} and show the results in Fig.~\ref{fig:rocresults}. 
It can be observed that the performances on Diffusion-based models including the LSGM and Latent-diffusion models and Transformers are lower than that on StyleGAN2, StyleGAN3, and 3DGAN models. This can be attributed to the fact that our detector is trained on images generated by a single GAN model and it can be extended to more general GAN model-generated image detection. However, it is challenging for the detectors to generalize well to the Diffusion models/Transformers due to the big gap in model structure and training strategy between GAN models and Transformer/Diffusion models. Similar results can be found in the baseline Nodown model. Overall, the results demonstrate that although the performance of our model is slightly lower than the Nodown model for unprocessed and anti-forensics images, our model achieves better performance on the other four protocols, which is consistent with that in Table~\ref{tb:test2}. 

\vspace{-0.4cm}

\subsection{Evaluation on Existing Datasets}
There are several existing datasets that consist of fake images from diverse generation models. In this section, we collect them to evaluate the generalization ability of the baselines and our method. Table~\ref{tb:test1} reports the average AUC on existing datasets and our collected dataset. More details of these datasets can be found in Table~\ref{tb:existingtestingdataset}. It can be seen that most methods achieve good performance on the testing dataset from the CNN-aug~\cite{wang2020cnn}. Our method obtains an AUC of 0.946. The testing data from ReGMs~\cite{asnani2021reverse} consists of more diverse generation models such that the performance on this dataset is slightly worse than that in CNN-aug for most detection methods. Our method achieves the best performance on this dataset with the help of combining global and local information. At last, to evaluate if current detection models are robust to the data generated by newly released models, we test all the methods on our collected dataset and report the AUC in the last row. It can be seen that all the methods drop greatly on this dataset, which demonstrates that the development of the generation model has made existing detectors more and more vulnerable, leading to more wrong predictions. Overall, our GLFF model achieves the best performance on most testing datasets. This demonstrates the outstanding generalization ability of our model benefiting from the fusing of global and local features.

 {As shown in Table~\ref{tb:test2} and Table~\ref{tb:test1}, on average, our method surpassed the baselines in three of the four test datasets. However, in certain subsets, such as Anti-forensics and Multi-image Compression, our model may have exhibited underperformance. This could be attributed to the fact that our model solely depends on spatial-domain information and may not take into account frequency clues that are essential for these subsets.}



\begin{table*}
\renewcommand\arraystretch{1}
\center
\footnotesize
\vspace{-0.1cm}

\caption{\label{tb:test1} \textbf{Evaluation results (AUC) on public datasets. The highest value is highlighted in black.}} 
\vspace{-0.2cm}

\scalebox{0.95}{
\begin{tabular}{ c | c c | c c |c c }
 \hline
 
   \textbf{Dataset} & \textbf{CNN-aug~\cite{wang2020cnn}} & \textbf{GAN-DCT~\cite{frank2020leveraging}} & \textbf{Nodown~\cite{gragnaniello2021gan}} & \textbf{BeyondtheSpectrum~\cite{yang_ijcai21}} & \textbf{PSM\cite{ju2022fusing}} & \textbf{GLFF~(Ours)} \\
   \hline
    \hline
   \textbf{CNN-aug~\cite{wang2020cnn}} & 0.936 & 0.603 & \textbf{0.968} & 0.875 & 0.942 &  0.946 \\
   \hline
   \textbf{ReGMs~\cite{asnani2021reverse}}  & 0.920 & 0.589 &  0.952 & 0.785 & 0.967  &\textbf{0.969} \\
   \hline
   \textbf{Newly-released models} & 0.569 & 0.646 & 0.653 & 0.564 & 0.641 & \textbf{0.665} \\
   \hline
\end{tabular}
}
\vspace{-0.4cm}

\end{table*}

\vspace{-0.4cm}

\subsection{Ablation Study}
To understand how different model architectures and module settings contribute to the performance of our proposed GLFF detection method, we compare our framework with various variants in terms of different model architectures, different module settings, and module parameters. We evaluate the variants and our method on the Protocol6-Mixed subset in our proposed $DF^3$ dataset considering that this subset can represent the general robustness to different manipulations.

\subsubsection{Architecture Variants}
In our work, we introduce a global and local feature fusion framework for  {AI-synthesized image detection}, in which the AMSFF and PSM modules are proposed to fuse multi-scale features and select patches, respectively. Here, we experiment with several alternative counterparts, \ie, (1) Global Branch Only: only utilizing the global branch in our two-branch model to detect forgery; (2) Local Branch Only: only the local embeddings are utilized for the final classification without being fused with global embedding $Em_g$; (3) Global~+~Local Branch w/o AMSFF: using the same two-branch structure as our GLFF, but removing the AMSFF module to fuse low-level and high-level features; (4) Global~+~Local Branch w/o PSM: using the same two-branch structure as our GLFF, but selecting patches randomly instead of using PSM. We present OA and AUC in Table~\ref{tb:abtest1}. It can be seen that only using the global branch or local branch achieves low performance while our GLFF model is the best in terms of both OA and AUC. This demonstrates that combining global and local features based on our proposed AMSFF and PSM can improve the generalization ability and robustness of our detection method.  {Global + Local Branch w/o PSM approach, which replaces the PSM module from the GLFF method with random patch selection, performed the poorest among all approaches. This results in a significant decrease in AUC performance of 0.160 when compared to our proposed GLFF model, dropping from 0.801 to 0.639. However, the Global + Local Branch w/o AMSFF, namely, removing the AMSFF module, achieved an AUC of 0.724. These findings suggest that the absence of the PSM module results in greater performance degradation and highlights the potential of the proposed PSM module to significantly enhance the detection performance. }


\begin{table}[!h]
\renewcommand\arraystretch{1}
\center
\vspace{-0.3cm}

\caption{\label{tb:abtest1} \textbf{Performance of our method based on different model architecture.}} 
\vspace{-0.2cm}

\scalebox{0.95}{
\begin{tabular}{ c | c | c | c | c | c | c | c}
 \hline
  \textbf{Model Structure}  & \textbf{OA} & \textbf{AUC} \\
\hline
\hline
Global Branch Only &  0.306 & 0.709 \\
  \hline
 Local Branch Only & 0.336 & 0.746 \\
  \hline
 Global + Local Branch w/o PSM  &  0.353 & 0.639 \\
 \hline
 Global + Local Branch w/o AMSFF & 0.313 & 0.724 \\
 \hline
 \hline
 GLFF (Ours)  & \textbf{0.355} & \textbf{0.801} \\
 \hline
\end{tabular}
}
\vspace{-0.2cm}

\end{table}

\subsubsection{Determination of Layers in AMSFF}
In the global branch of GLFF, we introduce the AMSFF module to combine features from shallow and deep layers based on an attention mechanism for feature extraction and patch selection. To understand how different layer combinations in AMSFF will contribute to the performance of our method, we choose different layer combinations in the AMSFF. 
There are 5 main stages in the structure of the ResNet-50 model, which are denoted from stages 1-5, respectively, as shown in Fig.~\ref{fig:resnet50}. As mentioned in~\cite{zhao2021multi}, subtle artifacts tend to be preserved in shallow layers of the network, thus we choose stage 1 and stage 2 as the candidates of the shallow layer. Conversely, we expect the detection method to grasp different regions of the input with the help of high-level semantic information. Therefore, we use stage 4 and stage 5 as the candidates for the deep layer. Experiment results in Table~\ref{tb:abtest2} demonstrate that the model achieves the best performance when using stage 1 as the shallow layer and stage 5 as the deep layer. 


\begin{table}[!h]
\renewcommand\arraystretch{1}
\center
\vspace{-0.2cm}

\caption{\label{tb:abtest2} \textbf{Performance of our method based on different combination of shallow layer and deep layer in AMSFF.}} 
\vspace{-0.2cm}

\scalebox{0.95}{
\begin{tabular}{ c | c | c | c | c | c | c | c}
 \hline
  \multicolumn{2}{c|}{\textbf{Candidates of Layers in AMSFF}}  & \multicolumn{2}{c|}{\textbf{Metric}}\\
  Shallow Layer  & Deep Layer  & OA & AUC \\
\hline
\hline
stage 1  & stage 4 & 0.298 & 0.759\\ 
  \hline
 stage 1 & stage 5 & \textbf{0.355} & \textbf{0.801} \\
\hline
stage 2 & stage 4 & 0.330 & 0.754 \\
 \hline
stage 2 & stage 5 & 0.332 & 0.693 \\
 \hline
\end{tabular}
}
\vspace{-0.1cm}

\end{table}

\vspace{-0.5cm}

\begin{table}[h]
\renewcommand\arraystretch{1}
\center
\caption{\label{tb:abtest3} \textbf{Performance of our method based on different size and number of window in PSM.}} 
\vspace{-0.2cm}

\scalebox{0.95}{
\begin{tabular}{ c | c | c | c | c | c | c | c}
 \hline
  \textbf{Window Size}  & \textbf{\# Window} & \textbf{OA} & \textbf{AUC} \\
\hline
\hline
$5 \times 5, 3 \times 3$ & 6 (3,3) & 0.330 & 0.779\\ 
  \hline
$3 \times 3, 2 \times 2$ & 6 (3,3) & \textbf{0.355} & \textbf{0.801}\\ 
  \hline
$3 \times 3, 2 \times 2$ & 4 (2,2) & 0.312 & 0.766 \\
\hline
$3 \times 3, 2 \times 2$ & 2 (1,1) & 0.346 & 0.749 \\
 \hline
\end{tabular}
}
\vspace{-0.3cm}

\end{table}

\subsubsection{Settings in PSM}
In our local branch of the proposed GLFF, we design the PSM module to select distinctive patches on the feature maps outputted from the global branch automatically. This module scans the feature maps with the window and selects the patches with higher energy, which contain more information for classification. 
During our implementation, the size and the number of windows are pre-defined. To understand how the window size and number of windows in PSM will contribute to the performance of our method, we compare the results in Table~\ref{tb:abtest3}. Since the combination of stages 1 and 5 achieves the best performance in AMSFF, the size of the fused feature map for patch selection is set as $7\times7$. Therefore, we experiment with several reasonable window sizes, \ie, $5\times5$, $3\times3$ and $2\times2$ and a different number of windows, \ie, 6 (3 per size), 4 (2 per size) and 2 (1 per size). Since more selected windows will lead to more computational resources, we set the maximum number of windows as 6 in our implementation. The experiment results in Table~\ref{tb:abtest3} show that the model reaches the best performance when scanning the feature maps with 3 windows with a size of $3\times3$ and 3 windows with a size of $2\times2$ in PSM.

\vspace{-0.3cm}
\subsection{Selected Patch Visualization}
\begin{figure}[t] 
\centering 
\vspace{-0.2cm}

\includegraphics[width=0.48\textwidth]{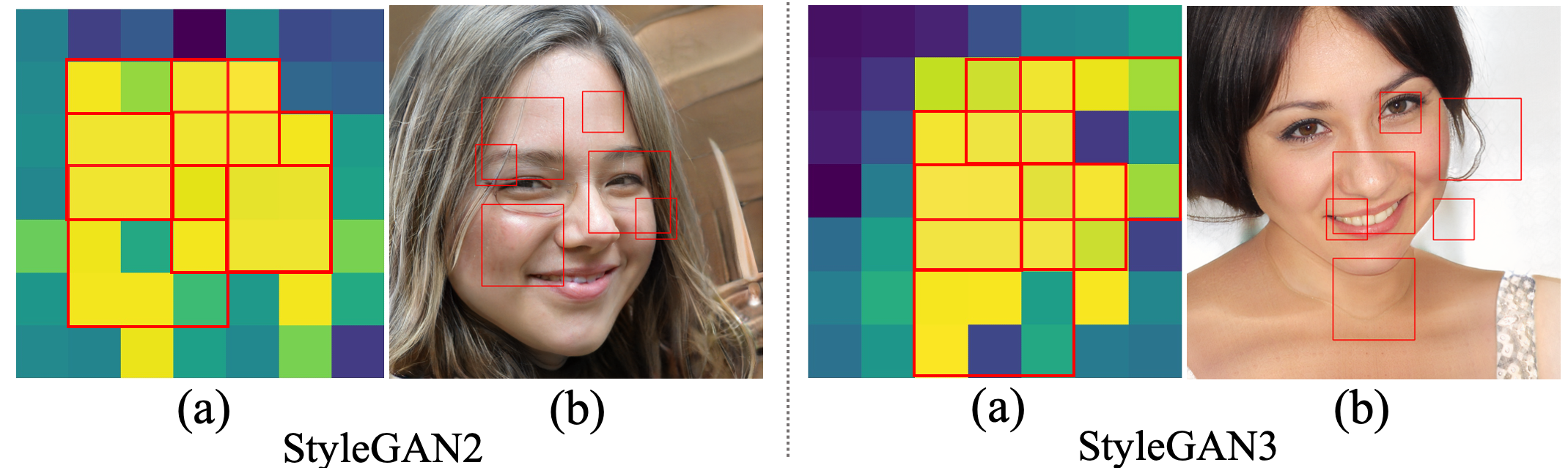}
\caption{\textbf{Visualization of selected patches on (a) Fused feature map, (b) Input RGB image. }} 
\vspace{-0.2cm}
\label{fig:patchvis} 
\end{figure}

To fully understand what patches Patch Selection Module~(PSM) focuses on, we visualize the selected patches on both fused feature maps $F_{fused}$ and the corresponding input RGB images in Fig.~\ref{fig:patchvis}. 
We show the patches on the images generated by StyleGAN2 and StyleGAN3. 
Note that the fused feature maps are averaged over the channel dimension and turned to the gray-scale image to show. The more yellow the color is, the stronger energy the region will show. As shown in Fig.~\ref{fig:patchvis}, when the selected patches on the fused feature map are mapped back to the input RGB images, the RGB patches show some semantic artifacts, such as the weird background artifacts, skin, hair, eyebrow, or wrinkles regions with rich texture information. 

\vspace{-0.3cm}

\subsection{Training Parameters}

 {To assess the time efficiency of our model training, we presented the number of trainable parameters for our GLFF model and two baseline models in Table~\ref{tb:parameters}. Despite having two branches and attention modules for multi-scale and global-local feature fusion, our approach does not result in a significant increase in the number of training parameters compared to the two baselines. The reason for this is that certain hyperparameters, such as the pre-defined window size and quantity in the PSM of the local branch, allow for the direct calculation of patch coordinates and cropping from the generated feature maps without the necessity for training.}

\section{Conclusion}
\label{sec:conclusion}

In this work, we propose a two-branch model GLFF that fuses multi-scale global features and informative local features for  {AI-synthesized image detection}. The global branch learns the multi-scale global feature using the proposed AMSFF module, based on which the local branch selects the informative patches using the proposed PSM module and extracts local features. The multi-head attention mechanism is finally used to fuse these complementary features for binary classification. Our model learns to focus on the most discriminative local patches with the help of the global high-level features in an unsupervised way, which allows us to fuse the local and global information more efficiently. To evaluate the performance of our model in the real-world scenario, we also develop a new dataset, $DF^3$, which combines several state-of-the-art generation models and various post-processings practiced in real-world applications. To our best knowledge, this is the first  {synthesized face image dataset} considering both generation models and rich post-processing. Extensive experiments on the proposed $DF^3$ and existing datasets demonstrate that the proposed model GLFF outperforms baselines and achieves high accuracy and generalization ability. Both the code and dataset details will be released on \url{https://github.com/littlejuyan/GLFF}.

In future work, we aim to examine a differentiable patch selection module, enabling automatic guidance for the size and quantity of local patches during model training. Additionally, we plan to expand our dataset by incorporating more advanced generation models and post-processing techniques prevalent in social media. Furthermore, it is valuable to develop a synthesized image detection model offering explainable performance. To this end, we plan to create a patch selection module that leverages additional prior knowledge constraints and provides a semantic explanation of the detection results.

\begin{table}[t]
\renewcommand\arraystretch{1}
\center
\vspace{-0.4cm}

\caption{\label{tb:parameters} \textbf{ {Number of trainable parameters of our method and baselines.}}}
\vspace{-0.2cm}

\scalebox{1.05}{
\begin{tabular}{ c | c  c  c }
 \hline
\textbf{Model}  & CNN-aug~\cite{wang2020cnn} & PSM~\cite{ju2022fusing} & GLFF~(Ours) \\
\hline
\# Parameters & 23.5M & 26.1M & 26.8M \\
  \hline
\end{tabular}
}
\vspace{-0.4cm}
\end{table}

\bibliographystyle{IEEEtran}
\bibliography{refs.bib}{}

 

\begin{IEEEbiography}[{\includegraphics[width=1in,height=1.5in,clip,keepaspectratio]{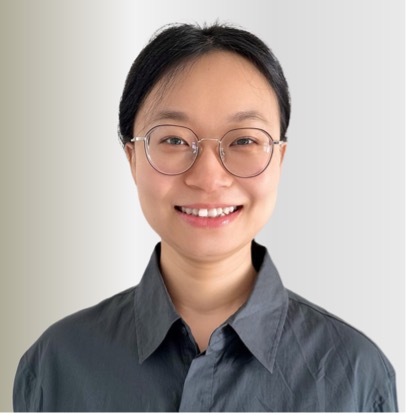}}]{Yan Ju} received the B.E. and M.E. degree from Xidian University, Xi’an, China, in 2016 and 2019. She is currently pursuing the Ph.D. degree with the Department of Computer Science and Engineering, State University of New York at Buffalo, Buffalo, New York, U.S. Her research interests include Fake Media Detection, Multi-media Forensics, and Computer Vision.
\end{IEEEbiography}

\vspace{-2cm}

\begin{IEEEbiography}[{\includegraphics[width=1in,height=1.5in,clip,keepaspectratio]{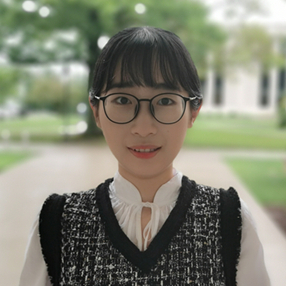}}]{Dr. Shan Jia} is a postdoc at the Department of Computer Science and Engineering of the University at Buffalo (UB), State University of New York, USA. She received her Ph.D. degree in Communication and Information Systems from Wuhan University, China, in 2021 and her B.S. degree in Electronic and Information Engineering from Wuhan University in 2014. Her research interests include multimedia forensics, biometrics, and computer vision.
\end{IEEEbiography}
\vspace{-2cm}

\begin{IEEEbiography}[{\includegraphics[width=1in,height=1.5in,clip,keepaspectratio]{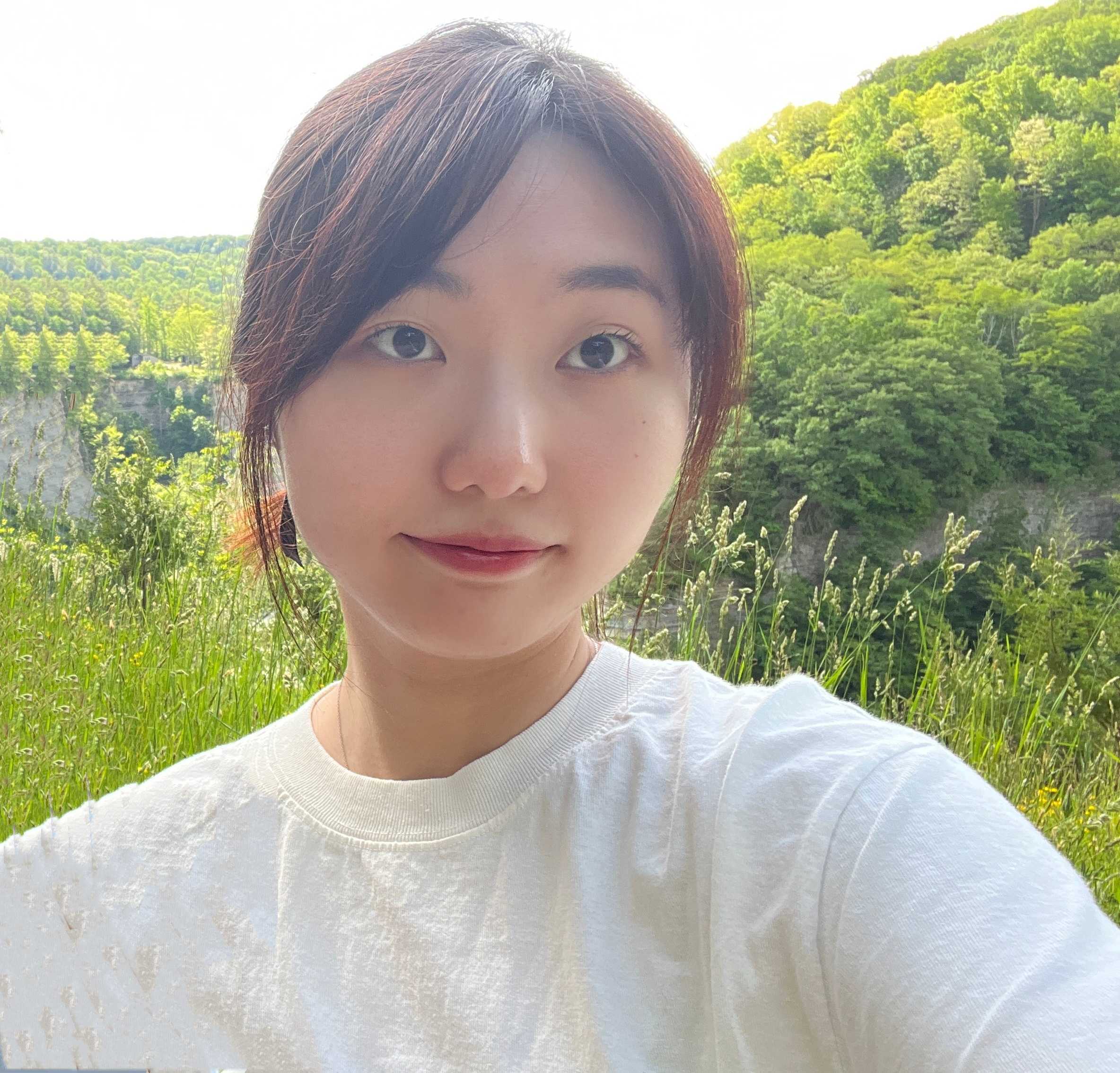}}]{Jialing Cai} is a first-year graduate student in the Department of Computer Science and Engineering at the University at Buffalo, State University of New York. She earned her Bachelor of Arts (B.A.) and Master of Science (M.S.) degrees in Econometrics from the University at Buffalo in 2022. Her research passion centers on media forensics and computer vision.
\end{IEEEbiography}

\vspace{-2cm}

\begin{IEEEbiography}[{\includegraphics[width=1in,height=1.5in,clip,keepaspectratio]{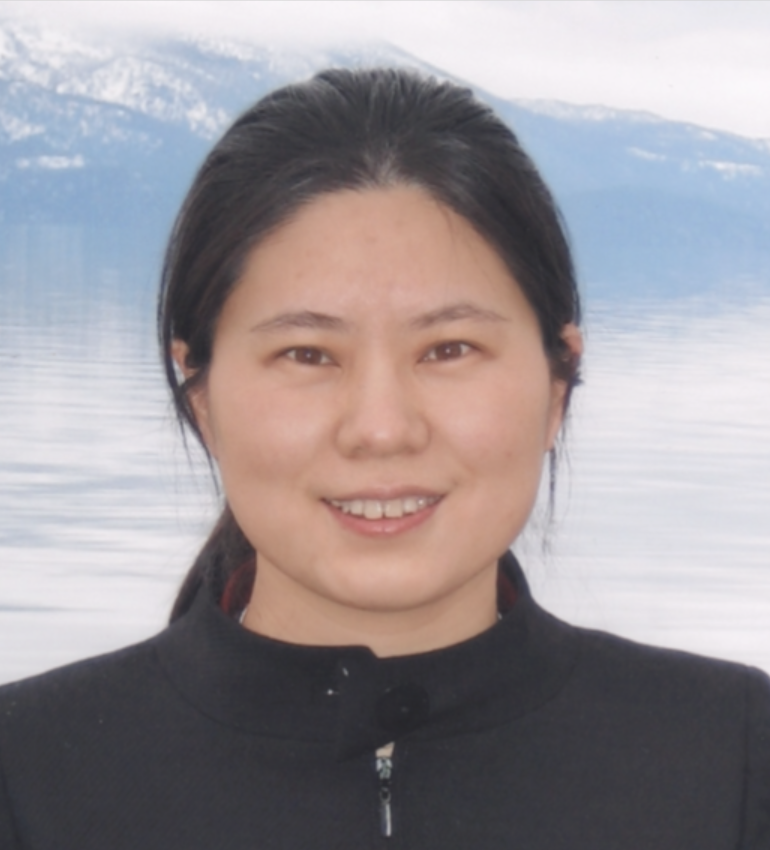}}]{Dr. Haiying Guan} is a Senior Computer Scientist at the National Institute of Standards and Technology (NIST) in the United States. She earned her Ph.D. and M.S. degrees in Computer Science from the University of California, Santa Barbara, in 2007. Dr. Guan currently serves as the lead for the Open Media Forensics Challenge (OpenMFC) evaluation program. She was a co-PI on the Media Forensics Challenge (MFC) evaluation project, which was sponsored by the Defense Advanced Research Projects Agency (DARPA) as part of the MediFor program. With a wealth of experience spanning government, industry, and academia, Dr. Guan has contributed to the fields of generative AI, media forensics, computer vision, biometrics, medical image processing, video analytics, quality measures, human-computer interaction, and usability. Her present research interests primarily revolve around the design of evaluation programs and the collection of benchmark datasets for the assessment of Artificial Intelligence (AI) systems.
\end{IEEEbiography}
\vspace{-1cm}
\begin{IEEEbiography}[{\includegraphics[width=1in,height=1.5in,clip,keepaspectratio]{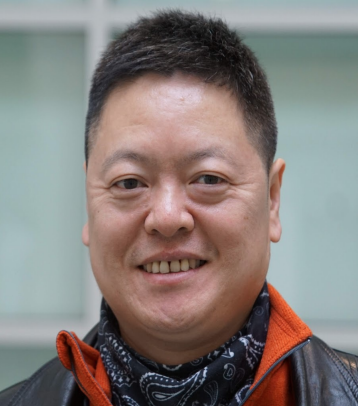}}]{Dr. Siwei Lyu} is a SUNY Empire Innovation Professor at the Department of Computer Science and Engineering, of the University at Buffalo, State University of New York, USA. Dr.
Lyu earned his Ph.D. in Computer Science from
Dartmouth College in 2005 and both his M.S.
(2000) and B.S. (1997) degrees in Computer
Science and Information Science, respectively,
from Peking University, China. Dr. Lyu’s research
interests include digital media forensics, computer vision, and machine learning. He is a Fellow of IEEE, IAPR, and AAIA, and a Distinguished Member of ACM.
\end{IEEEbiography}

\vfill

\end{document}